%% file: neurips_data_2022.tex
\pdfoutput=1
\documentclass{article}

\usepackage[numbers, compress]{natbib} 

\usepackage[preprint]{neurips_data_2023}
\usepackage{tablefootnote}
\usepackage{threeparttable}

\usepackage[utf8]{inputenc} 
\usepackage[T1]{fontenc}    
\usepackage{hyperref}       
\usepackage{url}            
\usepackage{booktabs}       
\usepackage{amsfonts}       
\usepackage{nicefrac}       
\usepackage{microtype}      
\usepackage{xcolor}         
\usepackage{textcomp}
\usepackage{graphicx}
\usepackage{multirow}
\usepackage{enumitem}
\usepackage{makecell}
\usepackage{ulem}
\usepackage{subfig}

\title{Fraud Dataset Benchmark and Applications}

\author{
	Prince Grover \\
	\texttt{pringrov@amazon.com} \\
	\And
	Julia Xu \\
	\texttt{amznxx@amazon.com} \\
	\And
    Justin Tittelfitz \\ 
    \texttt{tttlf@amazon.com}
    \And
	Anqi Cheng \\
	\texttt{anqichen@amazon.com}
    \And
	Zheng Li \\
	\texttt{mznzhen@amazon.com} \\
	\And
	Jakub Zablocki \\
	\texttt{jzablock@amazon.com} \\
	\And
	Jianbo Liu \\
	\texttt{lijianbo@amazon.com} \\
	\And
	Hao Zhou \\
	\texttt{zhuha@amazon.com} \\
}

\begin{document}

\maketitle
\begin{abstract}
  Standardized datasets and benchmarks have spurred innovations in computer vision, natural language processing, multi-modal and tabular settings. 
  We note that, as compared to other well researched fields, fraud detection has unique challenges: high-class imbalance, diverse feature types, frequently changing fraud patterns, and adversarial nature of the problem. 
  Due to these, the modeling approaches evaluated on datasets from other research fields may not work well for the fraud detection. In this paper, we introduce Fraud Dataset Benchmark (FDB), a compilation of publicly
  available datasets catered to fraud detection \footnote{Datasets are available at: https://github.com/amazon-research/fraud-dataset-benchmark}. FDB comprises variety of fraud related tasks, ranging from identifying fraudulent card-not-present transactions, 
  detecting bot attacks, classifying malicious URLs, estimating risk of loan default to content moderation. The Python based library for FDB provides a consistent API for 
  data loading with standardized training and testing splits. We demonstrate several applications of FDB that are of broad interest for fraud detection, including feature engineering, comparison of supervised learning algorithms, label noise removal,  class-imbalance treatment and semi-supervised learning. We hope that FDB provides a common playground for researchers and practitioners in the fraud detection domain to develop robust and customized machine learning techniques targeting various fraud use cases.  
  
\end{abstract}

\section{Introduction}\label{sec:intro}

Benchmark datasets of images \cite{imagenet, coco, mnist, krizhevsky2009learning}, videos \cite{soomro2012ucf101,
kay2017kinetics} and text \cite{wang2018glue, rajpurkar2016squad} have been extensively used in various research
publications and spurred the innovations in these fields. Standardized benchmarks help researchers disentangle the
impact of data and learning algorithms on the performance metrics. Common datasets ensure that the reported improvements
are independent of amount and diversity of data used in experiments, training/test splits or data distributions. By removing confounding factors related to data, standardized datasets allow direct comparison of pre-processing techniques and learning algorithms. Shared datasets also support
research reproducibility, without the need to publish production data. 

Classification and regression benchmarks concerning tabular datasets containing numeric, categorical and free form text features have been published in the past \cite{amlb2019, shi2021benchmarking, hu2020ogb, neuripsautomlchallenge, grinsztajn2022tree}. P. Gijsbers \cite{amlb2019} provides a benchmark of 18 publicly available multi-model tabular datasets for classification and regression tasks, and compares performance using various AutoML approaches. 
However, most of the datasets selected in that benchmark are not suitable for the study of fraud detection. Grinsztajn \cite{grinsztajn2022tree} intentionally removed stream-like datasets or time series which are not IID, balanced the class ratio in their benchmark datasets, and removed all high cardinality categorical features with more than 20 items.

We argue that the datasets required for fraud detection tasks have certain properties that make them distinctly different from currently available tabular data benchmarks. In particular many feature engineering and modeling techniques that work for other tabular datasets might not work well on fraud detection datasets. The unique properties of fraud datasets include:


\begin{enumerate}[topsep=0pt,itemsep=-1ex,partopsep=1ex,parsep=1ex]
 \item \textbf{Class imbalance}: The ratio of fraud to legit population can be extremely low (as low as 0.0001).
 \item \textbf{High cardinality of features}: Many raw data attributes in fraud space are highly granular (e.g. IP address, phone number). For those attributes, additional information can be extracted using 3rd party datasets (e.g. geo-location and internet service provider from IP address, mobile network from phone number, etc.).
  \item \textbf{Adversarial nature of the problem}: Fraudsters adapt their behaviors in order to fool the models and thus patterns in data are time dependent. As a direct consequence, the performance evaluations should be done ``out-of-time'' to accurately gauge the model's ability to work in practice.
  \item \textbf{Samples are not always Independent and Identically Distributed (IID)}: Attribute values and behaviors can depend on the historical values.
\end{enumerate}



Fraud detection is an important problem in both academic and industry research, with an impact 
that can prevent the loss of money, customer trust and privacy. Due to its importance and unique properties, we create the benchmark dedicated to fraud detection, with the goal of spurring the research in fraud detection and to facilitate the practical advancement in fraud detection. The FDB allows development of scientific approaches that are robust across multiple fraud MOs and makes it easy for researchers to share ideas.
All of the datasets in the benchmark are publicly available. We also provide a Python package \cite{fdb} that exposes a set of APIs to pull the datasets from their sources and the prepare the standardized train/test splits with common feature naming conventions. Each dataset can be loaded into a Pandas data-frame using the FDB data loader interface. Once a \texttt{FraudDatasetBenchmark} object is created with the required dataset key, the train
and test segments can be accessed through the class properties. The library allows one to optionally sample the data for easier data handling and  model training, and has methods to generate dummy variables required by some AutoML frameworks, like event timestamp, event ID etc. 
The source code to pull datasets from source and transform on the fly are released under MIT-0 license.

The rest of the paper is organized as the following: Section \ref{sec:datasets} gives a detailed introduction to the benchmark datasets and the package; Sections \ref{sec:super} -- \ref{sec:ssl} demonstrate four interesting applications of the benchmark: fraud detection modeled as binary classification using common ML algorithms as well as AutoML frameworks, label noise removal, class-imbalance treatment, and semi-supervised learning. Finally, Section \ref{sec:conclusion} concludes our findings and gives general guidance on the applicability of the benchmark in various business and technical scenarios, and ends with future work to improve the FDB.


\section{Benchmark Datasets}
\label{sec:datasets}

\subsection{Data Collection  and Selection}
\label{subsec:data_criteria}

In order to compile the benchmark, we searched various data sources including UC Irvine ML Repository
\cite{ucirvinemlp}, Kaggle \cite{kaggle}, Google dataset search \cite{googledatasearch}, Open ML \cite{openml}, UCR Time
Series Classification Archive \cite{UCRArchive2018}, The KONECT Project \cite{konect}, Stanford SNAP \cite{snapnets},
Papers with Code \cite{paperswithcode}, Awesome Public Datasets on Github \cite{awesomedata}, and previous AutoML papers
\cite{amlb2019}. 

FDB is currently limited to binary classification datasets (fraud vs. non-fraud). We selected 9 publicly available datasets from the fraud and abuse space. The datasets cover a range of common fraud problems including card not present transaction fraud, credit risk, bot attacks, and content moderation, all of which are suitable for comparing novel techniques such as label noise removal and semi-supervised learning specifically in the fraud domain. Out of the selected datasets, four of them (ieeecis, ccfraud, fraudecom, sparknov) have timestamps, three (ieeecis, fraudecom, sparknov) have user IDs, and two (fraudecom, ipblock) have raw IP addresses. Those datasets are well-suited for studying feature engineering based on account history and the role of data enrichments in fraud detection. Besides this, there are three datasets (ccfraud, ieeecis, fakejob) with extremely imbalanced class ratio (< 5\%), which are particularly useful for class imbalance studies. Following are further details about FDB datasets:



\begin{enumerate}
  \item \textbf{ieeecis} \cite{ieeecisdata}: Prepared by IEEE Computational Intelligence Society, this card-non-present transaction fraud
        dataset was launched during IEEE-CIS Fraud Detection Kaggle competition, and was provided by Vesta Corporation. The original dataset contains 393 features. We reduced the number of features to 67 features based on highly voted Kaggle kernels \cite{xgbfraudwithmagic,ieeefeatureselection}. 
        


  \item \textbf{ccfraud} \cite{ccfrauddata}: This dataset contains anonymized credit card transactions by European cardholders in September 2013. The dataset contains 492 frauds out of 284,807 transactions over 2 days. Data only contains numerical features that are the result of a PCA transformation, plus non transformed \textit{Time} and \textit{Amount}.

  \item \textbf{fraudecom} \cite{fraudecomdata}: This dataset contains 150k e-commerce transactions. The features include sign up time, purchase time, purchase value, device id, browser, and IP address. We added a new feature that measured the time difference between sign up and purchase, as the age of an account is often an important variable in fraud detection. 

  \item \textbf{sparkov} \cite{sparkovdata}: This is a simulated credit card transaction dataset. It was generated using Sparkov Data Generation tool \cite{sparkovgenerator} and we modified a version of dataset created for Kaggle. It covers transactions of 1000 customers with a pool of 800 merchants over 6 months. 

  \item \textbf{twitterbot} \cite{twitterbotdata}: The dataset is composed of 37,438 rows corresponding to different user accounts from Twitter. Each row contains features like account creation date, follower and following counts, profile description, account age, metadata about profile picture, account activity, and a label indicating whether the account is human or bot. 

  \item \textbf{malurl} \cite{malurldata}: This is a dataset of 650k URLs, among which 34.2\% of the URLs are malicious (e.g., used for defacement, phishing and malware). Even though the original dataset has multi-class labels we converted it into binary label.The kaggle dataset is curated using five different sources. 

  \item \textbf{fakejob} \cite{fakejobdata}: This Kaggle dataset contains 18k job descriptions out of which about 800 are fake. The data consists of both textual information and meta-information about the jobs. The task is to train a classification model to detect which job posts are fraudulent. 

  \item \textbf{vehicleloan} \cite{vehicleloandata}: The task is to predict the probability of vehicle loan default, particularly the risk of default on the first monthly installment. This dataset includes loanee information, loan information, credit bureau data, and history. 


  \item \textbf{ipblock}: In the fraud domain, 3rd party data enrichment of PII variables such as IP, email,
        and credit cards is a common practice and those features often have a large impact on performance. However, only 1 out of above 8 datasets had IP address (\textit{fraudecom}) due to data privacy concerns. Therefore, we created an additional dataset (\textit{ipblock}), made up from malicious IP address from CINS \cite{cins}. To the list of malicious IP addresses, we added randomly generated IP addresses using Faker \cite{faker} labeled as benign. 
        


\end{enumerate}


\begin{table}[!htbp]
\caption{Summary of the datasets in fraud dataset benchmark}
\label{tab:datasets}
\centering
\resizebox{\textwidth}{!}{%
\begin{threeparttable}
\begin{tabular}{@{}lllllllllll@{}}
\toprule
\textbf{Dataset name} & \textbf{Dataset key} & \textbf{Category}  & \textbf{\#Train} & \textbf{\#Test} & \begin{tabular}[c]{@{}l@{}}\textbf{Class ratio} \\ (train)\end{tabular} & \textbf{\#Feats} & \textbf{\#Cat} & \textbf{\#Num} & \textbf{\#Text} & \textbf{\#Enrichable} \\ \midrule 
IEEE-CIS Fraud Detection  & ieeecis & CNP & 561,013 & 28,527 & 3.50\% & 67 & 6 & 61 & 0 & 0 \\ 
Credit Card Fraud & ccfraud & CNP & 227,845 & 56,962 & 0.18\% & 28 & 0 & 28 & 0 & 0 \\ 
Fraud ecommerce & fraudecom & CNP & 120,889 & 30,223 & 10.60\% & 6 & 2 & 3 & 0 & 1 \\ 
Simulated CC Transactions & sparknov & CNP & 1,296,675 & 20,000 & 5.70\% & 17 & 10 & 6 & 1 & 0 \\ 
Twitter Bots & twitterbot & BA  & 29,950 & 7,488 & 33.10\% & 16 & 6 & 6 & 4 & 0 \\ 
Malicious URLs  & malurl & MT & 586,072 & 65,119 & 34.20\% & 2 & 0 & 1 & 1 & 0 \\ 
Fake Job Posting  & fakejob & CM & 14,304 & 3,576 & 4.70\% & 16 & 10 & 1 & 5 & 0 \\ 
Vehicle Loan Default & vehicleloan & CR & 186,523 & 46,631 & 21.60\% & 38 & 13 & 22 & 3 & 0 \\
IP Blocklist & ipblock & MT & 172,000 & 43,000 & 7\% & 1 & 0 & 0 & 0 & 1\\ 
\bottomrule
\end{tabular}
\begin{tablenotes}\footnotesize
\item[*] Category: CNP for Card-Not-Present Transaction Fraud; BA for Bot Attacks; MT for Malicious Traffic; CM for Content Moderation; CR for Credit Risk; MT for Malicious Traffic.
\end{tablenotes}
\end{threeparttable}
}

\end{table}

\subsection{Limitations}\label{subsec:limitations}

The datasets comprising FDB are, to the best of our knowledge, the most representative among publicly available datasets
of fraud detection problems. At the same time we are aware of challenges and limitations with
the selection we have made. First, we do not claim the FDB to be comprehensive, but hopefully with time the FDB
collection will grow to cover more business scenarios and dataset variations. Second, some datasets are missing quite important features such
as timestamp or user id that can be very useful to uncover time patterns and entity relations in the data. While we can
synthetically generate those fields, fraud detection algorithms that could in theory exploit the non-IID nature of the problem will not be able to actually benefit from these artificial features. Third, while these datasets are useful for research 
and development of fraud detection algorithms, they do not 
carry any information about real fraud. It someone uses any models trained on these datasets to directly make decisions about fraud, it would cause a negative bias and could lead to false accusations. Forth, the inclusion of personally identifiable information (PII) data, such as email address or
payment instrument information, is extremely rare. Only a single dataset has a full set of real IP addresses and no other PII
data elements are available. Unfortunately we can not expect this to change in the future, as there is justified growing public
concern about data privacy. Some of the datasets alleviate the problem of confidentiality through anonymization of PII
attributes through an irreversible set of data transformations. However, this has two consequences. First, the data enrichment techniques that rely on clear text become useless (e.g. extracting geolocation data from IP address) and so the researcher is dependent on data curator to perform those enrichments/transformation before data anonymization. Second, the anonymization process very often destroys the useful information present in data attributes (e.g. patterns in the email alias). From experience we know that those patterns are often critical to building highly performant solution in fraud detection.


\section{Application I: Data Enrichment and Feature Engineering in Supervised Learning}\label{sec:super}

Although in the real world, organizations hire teams of investigators to carefully review and label user's activities into an array of various classes of fraud and abuse, at the most fundamental level we can consider the problem of fraud detection as binary classification with individual events being either fraudulent (1) or legitimate (0). In that context we compare different supervised binary classification algorithms and AutoML frameworks. We compare commonly used algorithms from classification literature, including Random Forest \cite{rf}, Catboost \cite{cb}, LightGBM \cite{lgbm}, Multi Layer Perceptron (MLP) \cite{mlp}. We supplemented these baselines with state-of-the-art AutoML frameworks, which often optimize model hyperparameters or ensemble multiple models. In our study we included AutoGluon \cite{agtabular}, H2O \cite{H2OAutoML20}, Auto-sklearn \cite{feurer-arxiv20a} and Amazon Fraud Detector (AFD) \cite{afdtechguide} (Online Fraud Insight (OFI) and Transaction Fraud Insights (TFI)). Further details and comparisons of the chosen AutoML frameworks are discussed in Appendix \ref{appendix:auto_ml}.


In the fraud domain many features are not suitable for model training in their raw format and require careful data preprocessing. From our experience we often find that expert-crafted feature engineering can be critical for achieving strong model performance. Use of supervised machine learning and different feature engineering techniques for credit card fraud detection has been discussed in the literature \cite{AFRIYIE2023100163, DORNADULA2019631, 8316850}. \cite{8316850} proposes using behavioral patterns and sliding-window-based aggregations of transactions to train better models. \cite{RODRIGUES2022101207} evaluate the role of incorporating correlation and statistical inference between the current order and previous fraudulent ones as features. Some notably discussed features are the number of purchases of the user, the amount of each purchase, the number of tested credit cards, and the navigation pattern. We studied the impact of two feature engineering techniques: 1) use of data aggregates like count, age and recency of an activity of a user and 2) use of data enrichments like enriching IP address with 3rd party databases to extract additional information like geolocation. Feature engineering requires domain knowledge and time. Our goal with these baseline experiments is to provide a nudge to the researchers to propose new techniques in deriving useful model signals or automating the whole process. 

\begin{table}[!htbp]
\centering

\small

\caption{Benchmarking conventional feature engineering and modeling techniques. For this study, we used a subset of datasets from FDB that have at-least one of the 1) timestamp information from the source or 2) enrichable feature. 
Feature group R means raw features from FDB, R+A means raw as well as count, age and recency aggregates, R+A+E means 
raw, aggregates as well as IP enrichments. The enrichments include geolocation of an IP up to postal code granularity, internet service provider and autonomous system number of the IP. }

\label{tab:binary_class}
\scalebox{0.95}{
\begin{tabular}{cccccc}
\toprule
\textbf{Feature Group}         & \textbf{Model} & \multicolumn{4}{c}{\textbf{Dataset AUC-ROC}} \\ 
\midrule
                               &                & fraudecom  & ieeecis  & sparknov  & ipblock  \\ 
R                            & Random Forest  & 0.491      & \textbf{0.923}    & 0.929     & 0.5      \\
                               & Catboost       & 0.518      & \textbf{0.923}    & \textbf{0.995}     & 0.5      \\
                               & LightGBM       & 0.516      & 0.921    & 0.966     & 0.5      \\
                               & MLP            & 0.510      & 0.892    & 0.906     & 0.5      \\
\midrule
R+A               & Random Forest  & \textbf{0.646}      & 0.916    & 0.926     & 0.5      \\
                               & Catboost       & 0.638      & \textbf{0.923}    & 0.989     & 0.5      \\
                               & LightGBM       & 0.639      & 0.917    & 0.893     & 0.5      \\
                               & MLP            & 0.612      & 0.883    & 0.504     & 0.5      \\
\midrule
R+A+E & Random Forest  & 0.642      & N/A      & N/A       & 0.890    \\
                               & Catboost       & 0.630      & N/A      & N/A       & \textbf{0.916}    \\
                               & LightGBM       & 0.631      & N/A      & N/A       & 0.915    \\
                               & MLP            & 0.644      & N/A      & N/A       & 0.756    \\
\bottomrule
\end{tabular}%
}
\end{table}

In order to compare out of the box model performance, we kept the default settings for each modeling framework and evaluated all models on fixed test sets. The scripts to reproduce results and full ROC curves are present in our Github repository \cite{fdb}.  
All models on all datasets are evaluated using AUC-ROC as the performance metric. 

The first takeaway is that, in some scenarios, the fraud patterns can not be directly extracted from raw attributes without specialized feature engineering. We observe that on \textit{fraudecom}, without adding aggregates calculated from the prior history of the user, no modeling technique can perform better than random. On diving deep into this dataset, we observe that fraudsters generally have a higher number of transactions than legitimate people, but the baseline models do not extract this information by default. When we explicitly add user profiling features like “number of times the current user has performed an activity in the past”, the model starts to perform better than random (from 0.52 to 0.65 AUC; comparing raw vs. raw + aggregates for Catboost in Table \ref{sec:binary}). Among AutoML frameworks, only AFD's TFI calculates aggregates using timestamp, event ID and user ID and shows non random performance on \textit{fraudecom} in Table \ref{all_results}. 

Secondly, variable-specific feature enrichments can be very helpful, when applicable. 
In another dataset, \textit{ipblock}, no model is able to perform better than random unless we add 3rd party enrichments to the IP address. Using MaxMind \cite{maxmind} databases, we extracted information like country, city, postal code, ISP, ASN etc. 
for each IP address and added as additional features. The fraud detection capability significantly improves with such enrichments from 0.5 to 0.92 AUC (Tables \ref{tab:binary_class}). AFD's OFI yielded the highest AUC (0.94) with it's custom enrichments, but notably AutoGluon was able to achieve 0.81 AUC without IP enrichments. This re-emphasises the benefit of studying generalizable techniques for feature engineering over iterating on the modeling algorithms, and is often true when we deal with highly granular features like IP address, phone number, email address, etc. 

It is also evident that each dataset has a different level of information in distinguishing fraud and legitimate population. This is true in real scenarios too, with some fraud patterns being easy to detect, while other attacks are very hard to detect. Such differences also arise because different organizations have different levels of fraud detection capabilities. Some organizations might not even have simple rules/block-list based fraud prevention in place and fraudsters find it easy enough to enter into their systems, while other organizations might have advanced ML based fraud detection systems in which only advanced fraudsters can get into. The diversity in benchmarks will help us devise robust approaches for both basic and advanced fraud scenarios.


\begin{table}[!htbp]
  \centering
  \small
  
  \caption{Benchmarking AutoML frameworks.
    AFD TFI is targeted to cater transactional fraud use case and hence, it is only run for a subset of FDB. 
    The best performing model for each dataset is highlighted in \textbf{bold}.}
    
  \label{all_results}
\scalebox{0.95}{
    \begin{tabular}{cccccc}
      \toprule
      Dataset key  & \multicolumn{5}{c}{AUC-ROC}                                                                                                      \\
      \midrule
                   & AFD OFI                     & AFD TFI                               & AutoGluon      & H2O              & Auto-sklearn                  \\
      \midrule
      ccfraud      & 0.985                       & 0.99                                  & 0.99           & \textbf{0.992}   & 0.988            \\
      fakejob      & 0.987                       & -                                     & \textbf{0.998} & 0.99             & 0.983             \\
      fraudecom    & 0.519                       & \textbf{0.636}                        & 0.522          & 0.518            & 0.515             \\
      ieeecis      & 0.938                       & \textbf{0.94}                         & 0.855          & 0.89             & 0.932              \\
      malurl       & 0.985                       & -                                     & \textbf{0.998} & Training Failure          & 0.5        \\
      sparknov     & \textbf{0.998}              & -                                     & 0.997          & 0.997            & 0.995               \\
      twitterbot   & 0.934                       & -                                     & \textbf{0.943} & 0.938            & 0.936               \\
      vechicleloan & \textbf{0.673}              & -                                     & 0.669          & 0.67             & 0.664               \\
      ipblock      & \textbf{0.937}              & -                                     & 0.804          & Training Failure & 0.5             \\
      \bottomrule
    \end{tabular}%
  }
\end{table}

\section{Application II: Label Noise Removal}\label{sec:labelnoise}

\input{label_noise}

\section{Application III: Class Imbalance Treatment}\label{sec:classimbalance}
High class imbalance is a very common characteristic in the fraud detection domain, because for a normally operating business we expect the fraud rate to be reasonably low. There are various techniques in handling class imbalance such as 
\begin{enumerate}
    \item \textbf{Re-sampling techniques}: E.g. up-sampling the minority class, down-sampling the majority class, and Synthetic Minority Over-sampling Technique (SMOTE) which generate synthetic instances of the minority class by interpolating between existing instances \cite{chawla2002smote}; 
    \item \textbf{Cost-sensitive learning} assigning different mis-classification costs to the minority and majority classes, making the classifier more sensitive to the minority class \cite{elkan2001foundations}
    \item \textbf{Ensemble methods} combining ensemble learning with re-sampling, e.g., bagging with over/under sampling \cite{breiman1996bagging}, balanced random forest \cite{chen2004using} and RUSBoost \cite{seiffert2010rusboost}.
\end{enumerate}

In this section we select three datasets \textit{ccfraud, fakejob}, and \textit{ieeecis} whose fraud rate is below 5\% to compare the three re-sampling techniques and cost-sensitive learning via adjusting class weight in treating class imbalance in binary classification. We set three different levels of the {\it adjusted fraud rate} after the class imbalance treatment, which are 0.05, 0.1, and 0.5. For class weight we use the {\it scale\_pos\_weight} hyperparamter in CatBoost to match the weight of each class to the adjusted fraud rate. We continue to use AUC as our evaluation metric because it is robust against class imbalance.

\begin{figure}[!htbp]
    \centering
    \subfloat[]{\includegraphics[width=0.3\textwidth]{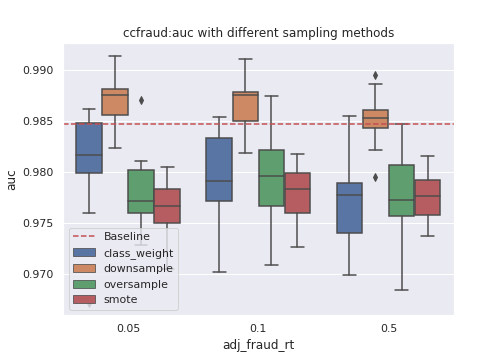}} 
    \subfloat[]{\includegraphics[width=0.3\textwidth]{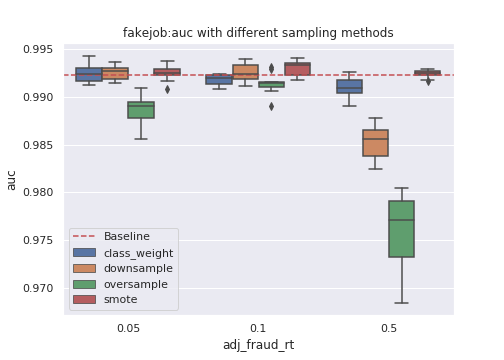}} 
    \subfloat[]{\includegraphics[width=0.3\textwidth]{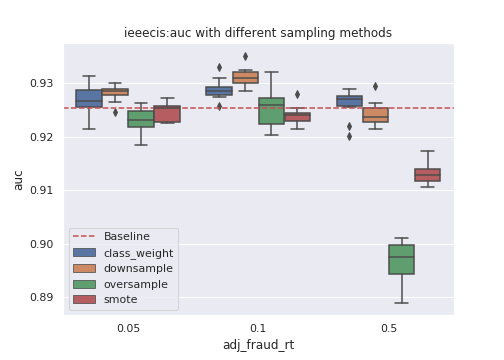}}
    \caption{Model performance (AUC) for different datasets after applying class imbalance treatment at three adjusted fraud rates.}
    \label{fig:imbalance}
\end{figure}

As we can see from the Fig \ref{fig:imbalance}, when the dataset size is large and fraud rate is low, appropriately downsampling the majority class is usually the most effective class imbalance treatment in improving model performance, while oversampling is generally discouraged. On the {\it ieeecis} dataset proper down-sampling can give 0.5\% AUC lift over the baseline AUC of 92\% and 0.25\% lift in AUC for {\it ccfraud} dataset with a baseline of 98.5\%. Caution is needed where the original data size is not very large, e.g., {\it fakejob} dataset is one order of magnitude smaller than the other two,  and aggressive down-sampling to achieve the high adjusted fraud rate of 0.5 makes the resulting dataset size too small and model performance suffers. We argue that the general effectiveness of down-sampling in imbalanced fraud datasets might correlate to the relative stable/predictable nature of legit population and violent/adversarial nature of the fraud population, such that in general reducing the majority/legit class in fraud domain is more effective in outstanding the minority/fraud class. 

\section{Application IV: Semisupervised Learning (SSL)}\label{sec:ssl}
In the struggle against fraudsters, labeling events can be challenging, costly and time consuming. In many cases, even if labeled data is scarce, unlabeled data may be readily available. Therefore semi-supervised learning, which utilizes both labeled and unlabeled examples, is of great interest both in theory and in practice. \cite{pise2008survey}.

Prediction accuracy is critical to fraud prevention, as even a small decline in model performance could result in significant monetary and reputational losses for an organization. However, there is no guarantee that incorporating unlabeled data in learning process will result in an increase of performance \cite{van2020survey, zhu2005semi}. Additionally, no algorithm has been found to consistently outperform the others \cite{chapelle2009semi, triguero2015self}. In this section, we use the benchmark datasets to evaluate and compare the effectiveness of various semi-supervised approaches in fraud detection against the supervised baseline.


\begin{enumerate}
\item \textbf{Supervised Baseline}: Oliver \cite{oliver2018realistic} mentions in his paper that the high-quality supervised baseline is critical. In many studies, the supervised baselines used for evaluating the semi-supervised learning methods are relatively weak, causing a biased expectation of the benefits of using unlabeled data. In our experiment, we use a cutting-edge boosting algorithm (Catboost) as the baseline, which can natively handle categorical features. 
\item \textbf{Wrapper methods}: Among the traditional inductive SSL methods, wrapper methods have outstanding flexibility since they are independent from the supervised algorithm and can be applied in conjunction with any supervised base learner. In this experiment, we evaluated two wrapper methods. a.) \textbf{Self-training } \cite{triguero2015self, amini2022self}, which iteratively retrains one classifier and adds confident pseudo-labeled data with a margin greater than a certain threshold or until no more data are left. b.) \textbf{Tri-training} \cite{zhou2005tri}, which starts with three classifiers trained on bootstrapped data, and then for each subset of data, pseudo-labeled samples agreed-on by the other two classifiers are added. This process is iterated but, to prevent overfitting, all pseudo-labels are recalculated, so that individual samples can be both added and removed from training. Beyond testing the default version, we also compared a conservative version of tri-training where we only add pseudo-labeled data if two classifiers strongly agree (both scores $>0.8$ for fraud, and $<0.2$ for legit).
\item  \textbf{Self- and semi-supervised frameworks}: Significant progress was made in the recent research on self- and semi-supervised learning techniques for multiple tasks. In many cases the approach can be seen as unsupervised pre-processing to learn representations, and also a way to incorporate unlabeled data in a second semi-supervised step through consistency or reconstruction loss. In this experiment, we add \textbf{VIME} \cite{yoon2020vime} as a benchmark comparison. Except for the default framework, we also test the strategy of only using the self-learning part to create additional features.
\end{enumerate}

Fraudulent activities tend to be clustered by fraud patterns, geo-locations, and time. Also, capturing new fraud patterns and allowing fraud labels to mature typically requires a certain amount of time. In this experiment, we simulate a scenario where early acquired events are more likely to be labeled and fraud events are detected by accounts. We first group all events by account ID. For the accounts with early acquired fraud, we label all their fraud events as fraud. For early acquired accounts with no fraud signals, we label all the events the from those accounts as legit. The rest events are considered as unlabeled. As in section \ref{sec:labelnoise}, we run this experiment for each of the 6 techniques on the 7 different datasets at 4 different fractions of labeled events (5\%, 10\%, 20\%, 50\%). For each dataset and labeling rate combination, we take note of the technique with the best performance, assigning it a rank. Results are shown in Table \ref{tab:ssl_rank}.

\begin{table}[!htbp]
\centering

\small

\caption{Rank of the test AUC with the 6 SSL technique at 4 different labeling rate (5\% - 50\%) }

\label{tab:ssl_rank}
\begin{tabular}{lllllll}
\toprule
\textbf{Rank} & \textbf{Baseline} & \textbf{SelfTrain} & \textbf{\begin{tabular}[c]{@{}l@{}}TriTrain \\ (Default)\end{tabular}} & \textbf{\begin{tabular}[c]{@{}l@{}}TriTrain \\ (Conservative)\end{tabular}} & \textbf{\begin{tabular}[c]{@{}l@{}}VIME \\ (Default)\end{tabular}} & \textbf{\begin{tabular}[c]{@{}l@{}}VIME\\ (Self + Catboost)\end{tabular}} \\ \midrule
\textbf{1st}  & 5 & 0 & 7 & 10 & 1 & 5 \\
\textbf{2nd} & 4 & 3 & 9 & 8 & 0 & 4 \\
\textbf{3rd}  & 12 & 0 & 9 & 7 & 0 & 0 \\
\textbf{4th}  & 4 & 10 & 1 & 3 & 1 & 9 \\
\textbf{5th}  & 3 & 14 & 2 & 0 & 2 & 7 \\
\textbf{6th} & 0 & 1 & 0 & 0 & 24 & 3 \\ \bottomrule
\end{tabular}
\end{table}

As shown in Table \ref{tab:ssl_rank}, the tri-training strategy works best on the fraud benchmark datasets, with the default version winning 7 times and conservative version ranking first 10 times. The self-training did not achieve the best performance in any scenario, which might be due to the its known overfitting problem. The default version of VIME framework did not perform very well, with only one first place on the fakejob dataset. As mentioned in Grinsztajn's work \cite{grinsztajn2022tree}, when compared to deep learning models (the semi-supervised part of VIME), tree-based models remain state-of-art on tabular data, which can be attributed the over-smoothing in NNs, many uninformative features in tabular data, and the rotationally non-invariant nature of the tabular data. In our experiments, we also added the representation learnt from the self-learning step as features into the CatBoost model. This approach took first place 5 times, but it's not significantly better than the baseline. In this work, we did not compare 

\section{Compute and Type of Resources}
\label{apx:compute}

AutoGluon, H2O and Auto-sklearn were trained using ml.c5.xlarge AWS Sagemaker notebook instance due to high compute requirements. AFD runs as a managed service within AWS and automatically infers the compute size. Unlike other frameworks we studied, AFD requires manual mapping of variable name to variable types \cite{afdvariable} and all of the variable mappings are available at our GitHub repository. All other experiments were performed on a GPU enabled p3.8xlarge EC2 instance. The Random Forest (scikit-learn), Catboost, LightGBM, and MLP (sckit-learn) were also used from respective official implementations with default hyperparameter settings. The exact details of experiment settings are also provided in the reproducibility folder in the repository.   

\section{Conclusions and Future Work}
\label{sec:conclusion}

We compiled and standardized publicly available datasets that are focused on fraud detection tasks, including card not present transaction fraud, bot attacks, malicious traffic, credit risk and content moderation. The standardized datasets can be loaded using Python with a simple APIs. 
We studied four interesting problems in fraud domain and demonstrate that a dedicated fraud data benchmark allows researchers working in the field of fraud detection to focus on designing and evaluating ML techniques instead of looking for appropriate training and testing data.



Some challenges like availability of PII data such as email addresses or payment information remain unmet in this benchmark.
At the same time, FDB is an on-going effort, and we hope to add more datasets and performance estimates in the future. Our
focus is on adding features that are hard to find publicly but have an impact in fraud detection. We will experiment
with designing anonymization techniques that can retain feature interaction, but also allow companies to share PII data.
Currently, we only benchmarked on 4 of the most interesting aspects of fraud detection problem. We plan to expand the performance evaluations on other applications for fraud detection like anomaly detection, graph neural networks, label propagation etc.



\input{references.tex}
\section*{Checklist}


\begin{enumerate}

\item For all authors...
\begin{enumerate}
  \item Do the main claims made in the abstract and introduction accurately reflect the paper's contributions and scope?
    \answerYes{Section 3 to 6 show four applications of this benchmark in fraud detection research. The README at Github repo \cite{fdb} provides sample codes to pull any dataset from this benchmark.}
  \item Did you describe the limitations of your work?
    \answerYes{See Section \ref{subsec:limitations}}
  \item Did you discuss any potential negative societal impacts of your work?
    \answerYes{In Section \ref{subsec:limitations}, limitation third, we discuss about potential bias if models trained on these datasets are used to directly make decisions about fraud.}
  \item Have you read the ethics review guidelines and ensured that your paper conforms to them?
    \answerYes{}
\end{enumerate}

\item If you are including theoretical results...
\begin{enumerate}
  \item Did you state the full set of assumptions of all theoretical results?
    \answerNA{}
	\item Did you include complete proofs of all theoretical results?
    \answerNA{}
\end{enumerate}

\item If you ran experiments (e.g. for benchmarks)...
\begin{enumerate}
  \item Did you include the code, data, and instructions needed to reproduce the main experimental results (either in the supplemental material or as a URL)?
    \answerYes{See scripts/reproducibility at Github repo \cite{fdb} }
  \item Did you specify all the training details (e.g., data splits, hyperparameters, how they were chosen)?
    \answerYes{See scripts/reproducibility at Github repo \cite{fdb} and Appendix \ref{appendix:auto_ml}.}
	\item Did you report error bars (e.g., with respect to the random seed after running experiments multiple times)?
    \answerYes{Error bars are reported where it is necessary for deriving conclusions, for example, Application III: Class Imbalance Treatment \ref{sec:classimbalance}. Bootstrapping and estimation of test variance is done for all other experiments (except AFD), but are not reported because of our reporting methodology. For example, in Application II and IV, we use ranking based report where multiple models can have same rank if the test scores are within the error bar. Is should be noted that, neither our experiments are meant to present statistically rigorous conclusions nor we claim a novel approach. We shared these four applications as motivating examples on how this benchmark can be useful for fraud detection research.}
	\item Did you include the total amount of compute and the type of resources used (e.g., type of GPUs, internal cluster, or cloud provider)?
    \answerYes{See Section \ref{apx:compute}.}
\end{enumerate}

\item If you are using existing assets (e.g., code, data, models) or curating/releasing new assets...
\begin{enumerate}
  \item If your work uses existing assets, did you cite the creators?
    \answerYes{See Data Sources section of the README page of Github repo.}
  \item Did you mention the license of the assets?
    \answerYes{See Section~\ref{sec:intro}, Github repo \cite{fdb} and supplemental material}
  \item Did you include any new assets either in the supplemental material or as a URL?
    \answerYes{Only one versioned dataset is uploaded, which is available at src/fdb/versioned\_datasets/ in the Github repo. All other datasets are generated on-the-fly.}
  \item Did you discuss whether and how consent was obtained from people whose data you're using/curating?
    \answerYes{We went through the license and intended use of each dataset and did not 
    include the ones that do not give consent.}
  \item Did you discuss whether the data you are using/curating contains personally identifiable information or offensive content?
    \answerYes{We discuss the absence of personally identifiable information and how it limits the scope of FDB benchmark in Limitations \ref{subsec:limitations}.}
\end{enumerate}

\item If you used crowdsourcing or conducted research with human subjects...
\begin{enumerate}
  \item Did you include the full text of instructions given to participants and screenshots, if applicable?
    \answerNA{}
  \item Did you describe any potential participant risks, with links to Institutional Review Board (IRB) approvals, if applicable?
    \answerNA{}
  \item Did you include the estimated hourly wage paid to participants and the total amount spent on participant compensation?
    \answerNA{}
\end{enumerate}

\end{enumerate}


\section*{Appendix}
\appendix

\section{Details of AutoML Frameworks}
\label{appendix:auto_ml}

\subsection{AutoML Frameworks}
\textbf{Amazon Fraud Detector (AFD)} \cite{afdtechguide} is a fully managed AutoML service that builds and productionalizes ML
models for fraud prevention. As per official documentation, it performs data validations, feature engineering and data
enrichments using open source, 3rd party, and Amazon knowledge base for specific variables commonly observed in fraud
detection, like IP address, email address, and card BIN. There are two supervised model templates in AFD, Online Fraud Insight (OFI) that handles general online fraud, and Transaction Fraud Insight (TFI) which extends OFI by introducing relational information between events and users from the past transaction activities.

\textbf{AutoGluon} \cite{agtabular} is an open source Python based AutoML toolkit developed by Amazon. It automatically
recognizes the data types of each feature, and discards features that are presumably of little predictive value (e.g.,
UserID). It preprocesses raw data based on data types (for example, it label encodes the categorical variables to reduce
memory footprint). It automatically takes care of missing values, and it splits raw data into training and validation
set when needed. It trains a variety of models including Random Forests, K Nearest Neighbors, LightGBM, CatBoost and
Neural Networks. It does not trigger hyper-parameter optimization by default. It instead uses a stacked ensemble of
different models or the same model with different hyper-parameters, and it uses repeated k-fold bagging to prevent
overfitting. For details on stacked ensemble and repeated k-fold bagging, see \cite{agtabular}.

\textbf{H2O} \cite{H2OAutoML20} is a Java based open-source library that offers AutoML capability. It is developed by the company
H2O.ai, which also sells a product called H2O Driverless AI built on top of H2O AutoML. H2O supports both Python and R
interfaces. It automatically preprocesses the raw data, which includes imputation, one-hot encoding and standardization.
The models in consideration include Gradient Boosted Machines, XGBoost, Deep Neural Networks, Generalized Linear Models,
etc. It uses random grid search for hyper-parameter optimization, with carefully chosen grids. It automatically tunes
individual models using cross validation. It employs stacked ensembles with two flavors: ``All Models'' and ``Best of
Family''.

\textbf{Auto-sklearn} \cite{feurer-neurips15a, feurer-arxiv20a} defines AutoML as a Combined Algorithm Selection and Hyperparameter optimization (CASH) problem. 
It is written in Python and leverages recent advantages in Bayesian optimization, meta-learning and ensemble construction. The
first paper \cite{feurer-neurips15a} was released in 2015 in which the authors introduced improvements in the Bayesian optimization framework by using meta-learning 
to warm-start the Bayesian optimizer and an automated ensemble construction step. The improvements led to an increased efficiency and robustness of AutoML. In 2020, 
Auto-sklearn 2.0 \cite{feurer-arxiv20a} was released, using meta-feature-free meta-learning technique and employing a successful 
bandit strategy for budget allocation.

In Table \ref{automl_comparisons}, we compare features supported by various AutoML frameworks and we hope it will be
helpful in deciding which framework meets the requirements for a given application. It's obvious that no single
framework is universally "the best" and careful considerations are needed to determine the winner.

\begin{table}[!htbp]
  \centering
  \caption{High level comparison of AutoML frameworks for fraud data compared in this work}
  \label{automl_comparisons}
  \resizebox{\textwidth}{!}{%
	    \begin{tabular}{lllllll}
		      \toprule
		      Pipeline                        & AFD OFI  & AFD TFI  & AutoGluon        & H2O              & Auto-sklearn                                                            \\
		      \midrule
		      Ensemble                        & N        & N        & Y                & Y                & Y                                                                         \\
          Meta Learning                   & N        & N        & N                & N                & Y                                                                          \\
		      3P enrichment                   & Y        & Y        & N                & N                & N                                                                              \\
		      Auto support for free form text & Y        & Y        & Y                & N                & N                                                                              \\
		      Aggregation features            & N        & Y        & N                & N                & N                                                                             \\
		      Graph features                  & N        & N        & N                & N                & N                                                                            \\
		      Fraud risk enrichment           & Y        & Y        & N                & N                & N                                                                         \\
		      Decision rules                  & Y        & Y        & N                & N                & N                                                                           \\
		      Blocklist                       & N        & N        & N                & N                & N                                                                           \\
		      Auto train/valid split          & Y        & Y        & Y                & N                & Y                                                                            \\
		      Cross validation                & N        & N        & Y                & Y                & Y                                                                       \\
		      Auto variable mapping           & N        & N        & Y                & Y                & Y                                                                            \\
		      Model explainability            & Y        & Y        & Y                & Y                & Y                                                                           \\
		      Auto scalable infrastructure    & Y        & Y        & N                & N                & N                                                                           \\
		      \bottomrule
		    \end{tabular}%
	  }
\end{table}

\subsection{Comparison of AutoML frameworks on FDB}
\subsubsection{High cardinality of features}

Raw attributes in typical fraud detection scenarios are often highly granular (e.g., IP address, phone, email address,
billing/shipping address, credit card number, device fingerprint). Treating such high cardinality attributes simply as
categorical variables is generally not useful for machine/deep learning models to predict the riskiness of fraud. Most of 
the algorithms in the tabular data domain encode categorical data (e.g. target encoding, one hot encoding, embeddings) such that same or 
similar categories lie in common buckets. Estimating the changes of fraud using buckets of categories leads to a 
high variance in estimates if the count of events per bucket is low (common in high cardinality cases) or in the worst case, 
makes it impossible to get an estimate due to unseen values in the training data.

There are two common modeling approaches that are commonly used in fraud domain to handle such high cardinality variables: 1)
extract relationships (e.g., whether the IP address has been used to commit fraud), 2) enrich using 3rd party solutions
(e.g. geo-location and internet service provider from IP address).

\begin{figure*}[!htbp]
  \centering
  \includegraphics[width=0.7\textwidth]{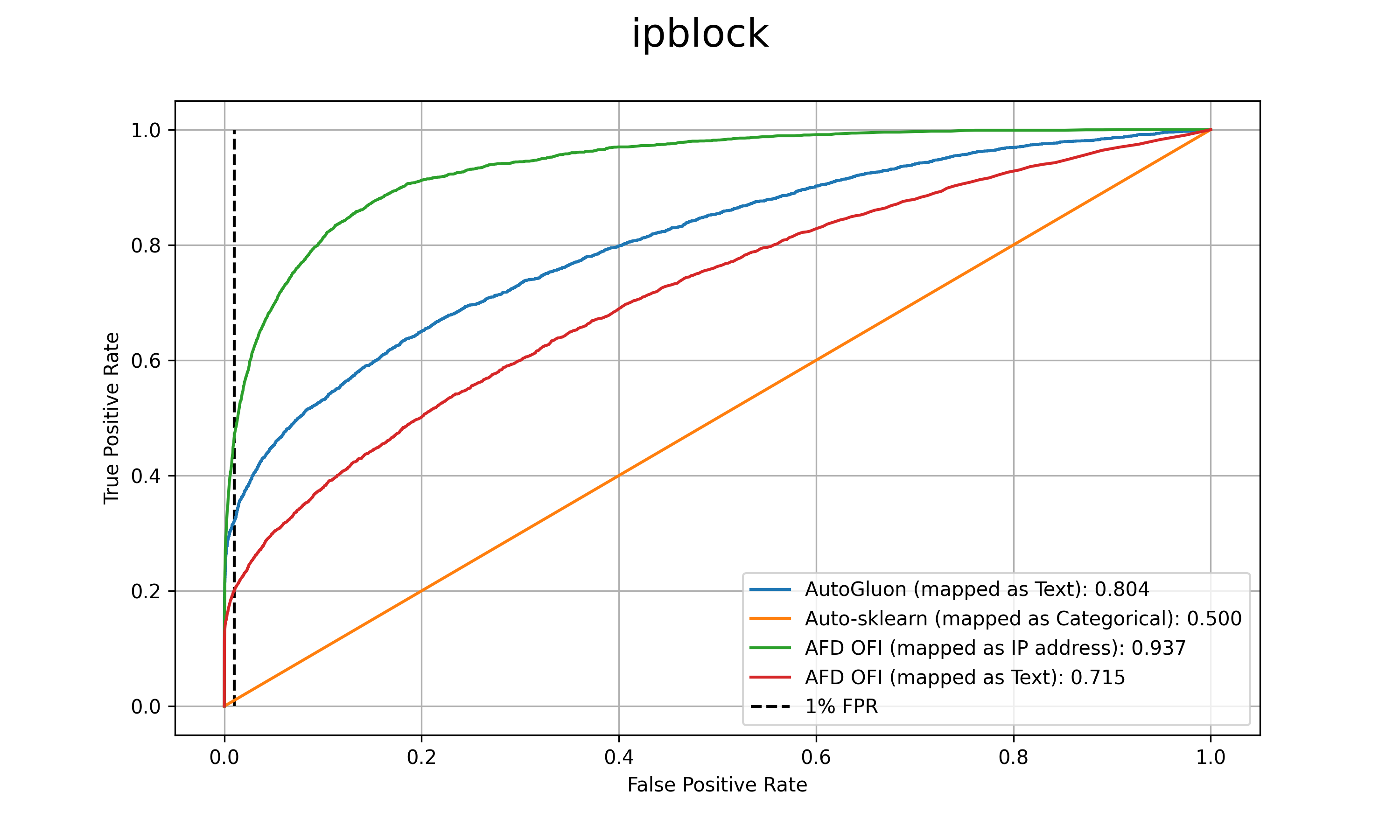}
  \caption{ROC curve for \textit{ipblock} dataset with different settings of AutoML frameworks.}
  \label{fig:roc_curves_fdb}
\end{figure*}

\textit{ipblock} contains real malicious IP addresses and are extremely granular at raw level, with 100\% uniqueness.
Since no other useful feature exists in this dataset, it is an ideal dataset to study a framework\textquotesingle s
capabilities in handling granular attributes. Table \ref{ipblock-table} and Figure \ref{fig:roc_curves_fdb} show
comparisons on this dataset. AutoGluon and H2O with default settings fail to correctly infer the variable type and treat
IP address values as Categorical. 100\% uniqueness leads to validation failures and both fail to train a model. 
Auto-sklearn does not have a validation check prior to the modeling and ends up training a random classifier using the 100\% unique Categorical feature.   
H2O and Auto-sklearn do not support Text processing, but one can force the IP address to be treated as Text in AutoGluon.


\begin{table}[!htb]
  \centering
  \caption{Test AUCs from different settings of variable mappings on \textit{ipblock} data}
  \label{ipblock-table}
  \resizebox{\textwidth}{!}{%
    \begin{tabular}{lllll}
      \toprule
      \textbf{Model Setting}           & \textbf{AFD OFI} & \textbf{AutoGluon} & \textbf{H2O}      & \textbf{Auto-sklearn}     \\
      \midrule
      Auto variable mapping            & Not supported    & Training failure   & Training failure  & 0.5                       \\
      IP address mapped as Categorical & Training failure & Training failure   & Training failure  & 0.5                       \\
      IP address mapped as Text        & 0.715            & 0.804              & Not supported     & Not supported             \\
      IP address mapped as IP address  & 0.937            & Not supported      & Not supported     & Not supported             \\
      \bottomrule
    \end{tabular}%
  }
\end{table}

AFD by nature requires variable mapping (i.e. does not support auto variable mapping), and supports Categorical, Text as
well as dedicated IP address. We try all three variable mapping combinations with OFI. Similar to other 2 frameworks, if
this variable is treated as Categorical, OFI fails to train the model. Text mapping performs worse than AutoGluon (0.715
vs. 0.804 AUC). Rightfully treating this variable as IP address leads to a substantial improvement over treating as Text
(0.715 vs. 0.937 AUC; shown in Figure \ref{fig:roc_curves_fdb}) emphasizing the value of feature enrichment in situations
encountered in practical fraud detection tasks.

\subsubsection{Adversarial nature of problem}\label{subsubsec:adversarial}

The adversarial nature of the fraud and frequently changing fraud patterns are generally solved using re-training model weights as the new data comes in. 
Therefore, it is important to incorporate the model deterioration when reporting the performance on unseen datasets. Otherwise, it leads to overestimation of 
estimated performance vs. what one can expect in production, leading to inaccurate business metrics. The expected performance is estimated using a local validation 
approach, e.g. cross validation or using local training/validation/test splits. Different local validation approaches can lead to different amount of 
reporting bias. 

An ideal model would have the least bias in the performance reported from the local validation compared to the
performance on out of time hold out set. Analyzing this behavior (Table \ref{local-validation} ) in datasets where
actual event timestamp is available and is not random (\textit{ccfraud, ieeecis}), the bias is high with AutoGluon and H2O (11.4\% and
6.3\% in AUC respectively) on \textit{ieeecis}. This is likely because the k-fold bagging and cross validation are not
done out-of-time. On \textit{ccfraud}, the bias is low in all models. AFD OFI and TFI capture adversarial nature of
fraud in split strategy by relying on out-of-time split strategy and provide a low bias in local validation vs. holdout test.

\begin{table}[!htb]
  \centering
  
  \caption{Difference in AUC from Local Validation (LV) and Holdout Test (HT) on $ccfraud$ and $ieeecis$.
          The number in \textbf{bold} highlights the model with lowest average bias betwen local validation and holdout test performance.}
  \label{local-validation}
    \begin{tabular}{cccccccc}
      \toprule
                                                          &
      \multicolumn{2}{c}{\textbf{LV}} &
      \multicolumn{2}{c}{\textbf{HT}} &
      \multicolumn{3}{c}{\textbf{|LV - HT|}}                                                                                                                                                        \\
      \midrule
                         & ccfraud & ieeecis & ccfraud & ieeecis & ccfraud & ieeecis & Average  \\ 
      \midrule
      AFD OFI            & 0.99    & 0.95    & 0.985   & 0.938   & 0.005   & 0.012   & \textbf{0.0085}    \\                     
      AFD TFI            & 0.98    & 0.95    & 0.99    & 0.94    & 0.01    & 0.01    & 0.01         \\
      AutoGluon          & 0.982   & 0.969   & 0.99    & 0.855   & 0.008   & 0.114   & 0.061        \\
      H2O                & 0.977   & 0.953   & 0.992   & 0.89    & 0.015   & 0.063   & 0.039        \\
      Auto-sklearn       & 0.977   & 0.947   & 0.988   & 0.932   & 0.011   & 0.015   & 0.013        \\
      \bottomrule
    \end{tabular}%
  
\end{table}

\subsubsection{Data samples are not IID}

Another differentiating property in fraud applications is that the attributes and behaviors of the new events depend on the past
information (non IID). For example, if a user logs in to a personal account once every day, but suddenly logins for the same user start 
coming once every minute, there are chances that the user's account is compromised. In such a scenario, 
a \textit{data aggregate} that calculates the \textit{count of logins} for a given user within 1 day from the current login time 
would be able to capture this information and learn this pattern even with simple tree based models. Similarly, for the detection 
of transactional fraud, features like user\textquotesingle s recent purchases (time and amount), time since account is created etc. 
play a vital role in powerful fraud detection. If such feature engineering is not performed, the tree based models would not be able 
to capture information about frequency of logins from the raw timestamps. 


Only two datasets (\textit{ieeecis} and \textit{fraudecom}) in FDB have both meaningful event timestamp and user id, and
only AFD TFI framework \cite{afdtechguide} uses aggregates in feature engineering, that are calculated using event
timestamp and user id . For both of these datasets, AFD TFI shows performance lift (in AUC) over other AutoML
frameworks, as highlighted in Table \ref{all_results}. The power of aggregates is also discussed in a notebook shared by
the competition winner of IEEE-CIS Fraud Detection competition \cite{afdtechguide}.

\subsubsection{Performance dependency on variable types}

The datasets in FDB contain different proportions of variable types in Numeric, Text, Categorial and Enrichable
categories. For the dataset that only contains Numeric features (\textit{ccfraud} in Table \ref{all_results}), the
difference in model performance in all frameworks is minor, showing an equivalent efficacy in handling Numeric features
among all frameworks. For the datasets containing a good proportion of Text features (\textit{twitterbot, fakejob,
malurl} in Table \ref{all_results}), AutoGluon consistently performs best in AUC. Both AutoGluon (default)
\cite{amlb2019} and AFD \cite{afdvariable} use n-gram and term frequency variant for Text processing. N-grams can be
generated by different tokenization procedures, and the tokenization, number of tokens and post-processing (if any) have
an impact on final model performance and are likely the reason for better performance with AutoGluon. H2O and Auto-sklearn 
are not designed to handle text features and automatically treat some columns considered as text as categorical instead (reason of
training failure on \textit{malurl} with H2O and random classifier with Auto-sklearn). Training failed for H2O as the pipeline drops 
irrelevant and unique categorical variables before starting the model training). 
Lastly, \textit{vehicleloan, fakejob and sparknov}, have a higher proportion of Categorical variables, where no single framework wins over others.



\subsubsection{Limitations in all AutoML frameworks}

One of the goals with a public fraud dataset benchmark is to help introduce new ideas in ML research by highlighting the
shortcomings in existing solutions. The experiments with AutoML frameworks on FDB datasets call attention to limitations
that currently exist in those frameworks. First, prior to training a model, AutoML should be able to robustly infer the
correct variable type (e.g. Text vs. Categorical) that determines the further processing steps. AFD does not support
variable type inference, and both H2O and AutoGluon fail to infer correct variable types in certain datasets including,
for example \textit{malurl}. Incorrect variable type mappings substantially decay the model performance as shown in Table
\ref{ipblock-table}. Secondly, none of the frameworks currently use graph based relational information to improve fraud
detection, especially for highly granular raw data that does not reveal information in standard tabular format. Similar
to relational variables, aggregation features based on historical information can not only improve the model
performance, but also make models stable over time (discussed in Kaggle kernel \cite{xgbfraudwithmagic}). Only AFD TFI
\cite{afdtechguide} currently extracts this information. Aggregates are challenging to implement in general purpose
AutoML frameworks as they require backend infrastructure that stores and updates its state as new inference samples are
scored. It's no surprise then that only TFI, a cloud based solution, can support that feature.

\section{Details of Label Noise Experiments}
\input{label_noise_appendix}

\end{document}

%% file: label_noise.tex
In this section we compare label noise removal techniques using the benchmark datasets. This problem arises frequently in fraud detection settings where observed labels in training data may differ from true labels due to adversarial behavior, label maturity issues, etc. and can negatively affect training, validation, and benchmarking of models. Recently there have been multiple advances in this area, but in order to assess the performance of these techniques, one needs to either clean noisy data (which requires expertise and is expensive) or to add realistic noise to clean data. We suggest that the benchmark datasets provide a good starting point for the second approach.

Here, we hope to demonstrate to the reader that these datasets are suitable for comparison of novel noise removal techniques to the existing state of the art. In particular, after adding noise (at various levels) to these datasets we see noise removal techniques outperform a naive baseline (where no noise removal is attempted). We also expect to identify the best performing method across all scenarios, or at least a class of scenarios, e.g. in high-noise situations, on datasets where fraud rate is low, etc. Through these, a practitioner may learn something about the relative strengths and weaknesses of their approach.


We think feature-dependent (vs. random or class-dependent) noise is the most realistic in application and so we generated "boundary-consistent" noise for our experiments, where we first trained a classifier on clean data, and then use distance to the decision boundary in order to weight selection of examples to mislabel (see \cite{menon2018learning}). This was implemented in Scikit-clean (see \cite{skclean}). In all cases we used CatBoost as the base classifier (see \cite{catboost}), as it works well out-of-the-box on tabular data without much need for hyperparameter tuning. In view of this choice, we omitted the \textit{ipblock} and \textit{malurl} datasets in this study as they are unstructured, and we dropped text and enrichable features from other datasets. For categorical features, we used target encoding (\textit{after} noise is added). For each experiment, we added boundary consistent noise to the training data, did the target encoding, attempted to remove noise using one of the techniques, trained a model on the cleaned data, and then evaluated this model on (clean) test data, taking note of the AUC. The techniques we used were as follows:

\begin{enumerate}
\item \textbf{Baseline}: No cleaning is done here, the model is trained directly on the noisy data.
\item \textbf{CleanLab}: This is a state-of-the-art technique that is implemented in an open-source package, see \cite{northcutt2021confident}. The main idea is to determine a threshold on model scores for each class, beyond which an example is discarded as noise. This package is compatible with any model that implements the sklearn interface.
\item \textbf{Scikit-clean MCS}: This is an open-source toolkit for noise generation and filtration. We choose the sequential Markov Chain Monte Carlo method (MCS) from this package (see \cite{zhao2019classification}). 
\item \textbf{Micro-models (MM)}: This approach trains an ensemble of models on disjoint slices of training data, and then lets the ensemble vote on whether each example is mislabelled. We experiment with majority and consensus voting schemes. See \cite{cretu2008casting}, \cite{samami2020mixed}, \cite{sabzevari2018two} for examples in the literature. 
\end{enumerate}


We ran this experiment for each of the five techniques on seven different datasets at six different noise levels (at levels 0, 0.1, 0.2, 0.3, 0.4, 0.5). We repeated it five times for each noise level and noise removal technique and computed the mean AUC and standard deviation. For each dataset/noise-level combination, we ranked the techniques based on the performance, assigning 1st, 2nd, etc. places, and allowing ties for results within a standard deviation. Results are shown in Table \ref{tab:label_noise_results}. Note that, at the noise level of 0, we expect Baseline to win. In this contest, CleanLab was a clear winner, with 18 first place finishes. The Micro-model Majority and skclean-MCS schemes each had been the best in 14 scenarios, Baseline ranked first 10 time, and Micro-model Consensus only 7. It is also notable that CleanLab was never the worst performing technique.

In the appendix, we show detailed performance across these experiments. In summary, Baseline usually performs well (as expected) with zero noise. Overall model performance declines for all methods as noise is added but the noise removal methods mitigate this effect to some extent. CleanLab performs very well across the wide range of scenarios, though at high noise levels MM-Majority and skclean-MCS sometimes perform better.

\input{tables/label_noise_race_results}

%% file: tables/label_noise_race_results.tex
\begin{table}
\centering

\small

\caption{Label Noise Removal Summary. The table lists the number of scenarios for each technique to rank at 1st, 2nd, ..., 5th place. Totals are not evenly distrubuted due to ties.}

\label{tab:label_noise_results}
\begin{tabular}{lrrrrrr}
\toprule
{} &  \textbf{Baseline} &  \textbf{skclean MCS} &  \textbf{CleanLab} &  \textbf{MM Majority} &  \textbf{MM Consensus} &  \textbf{Totals} \\
\midrule
\textbf{1st} &        10 &           14 &        18 &                    14 &                      7 &      63 \\
\textbf{2nd} &         7 &            6 &         7 &                    11 &                      8 &      39 \\
\textbf{3rd} &         4 &            5 &        11 &                     5 &                     16 &      41 \\
\textbf{4th} &         5 &           16 &         6 &                     7 &                      9 &      43 \\
\textbf{5th} &        16 &            1 &         0 &                     5 &                      2 &      24 \\
\bottomrule
\end{tabular}
\end{table}

%% file: label_noise_appendix.tex

In this section we provide more complete results for our experiments with label noise removal in Section \ref{sec:labelnoise}.

\begin{figure*}[ht]
  \centering
  
  \subfloat{\includegraphics[width=0.35\textwidth]{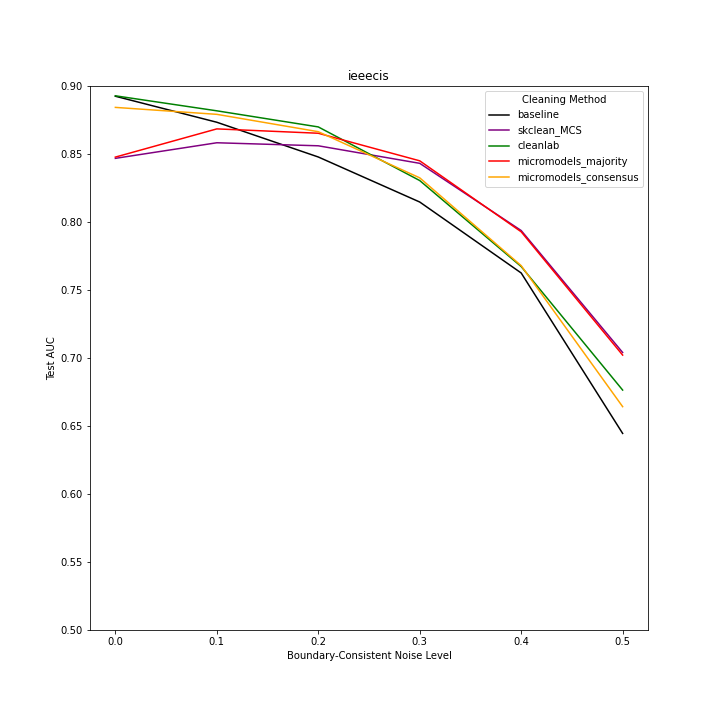}}
  \subfloat{\includegraphics[width=0.35\textwidth]{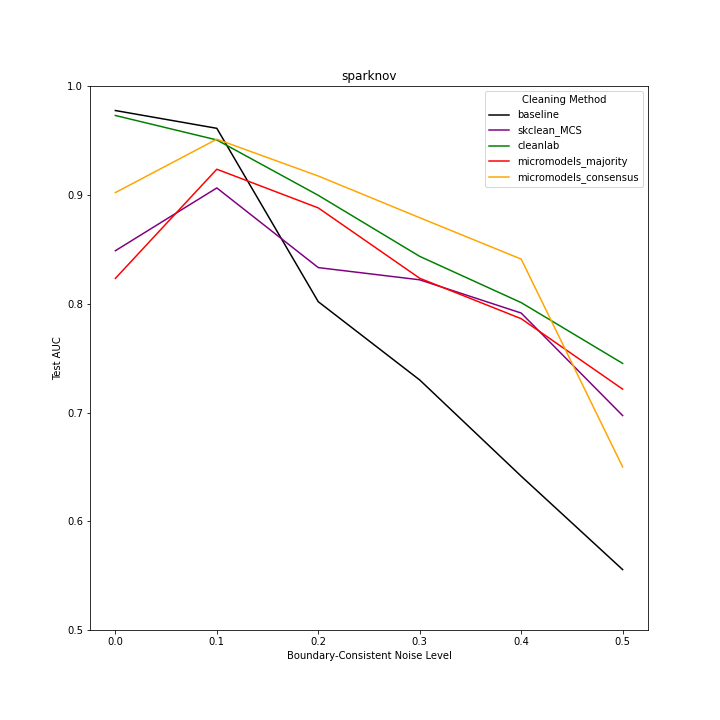}}
  \subfloat{\includegraphics[width=0.35\textwidth]{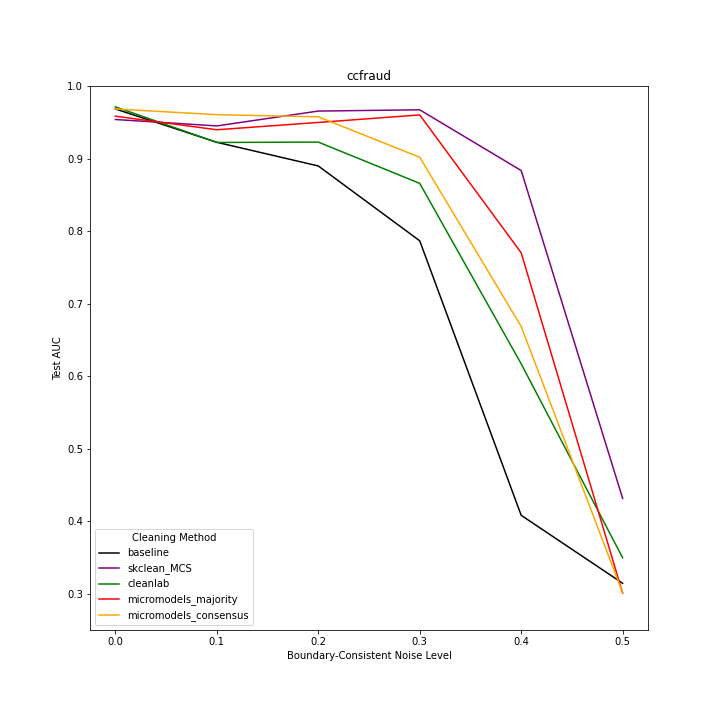}}\\
  \subfloat{\includegraphics[width=0.35\textwidth]{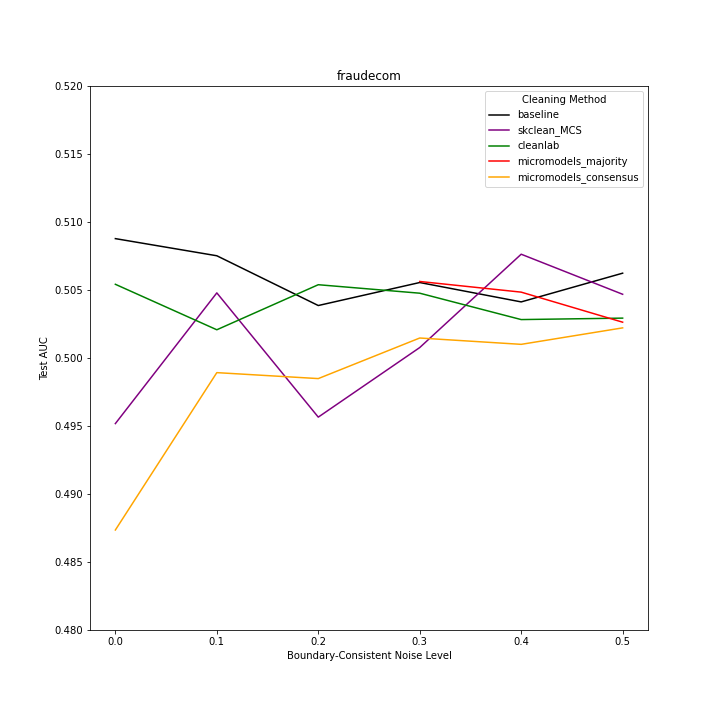}}
  \subfloat{\includegraphics[width=0.35\textwidth]{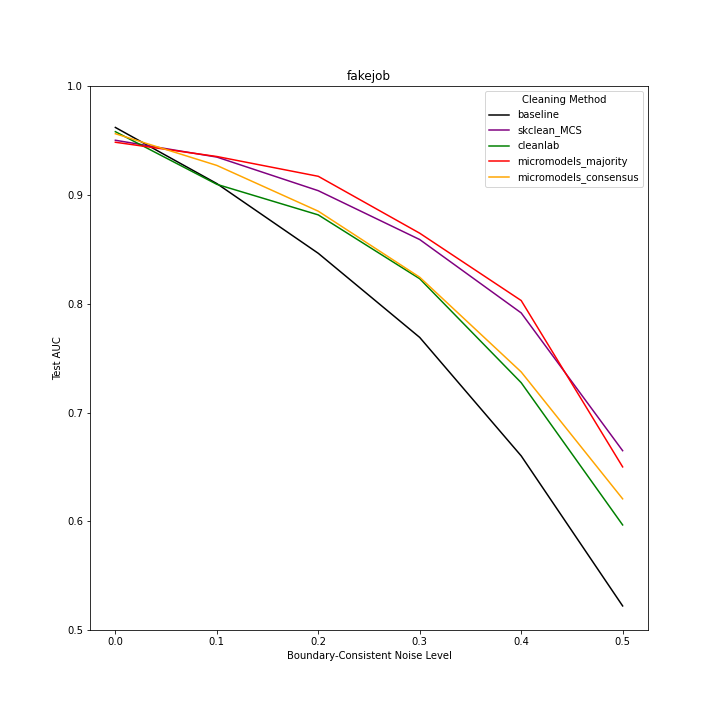}}
  \subfloat{\includegraphics[width=0.35\textwidth]{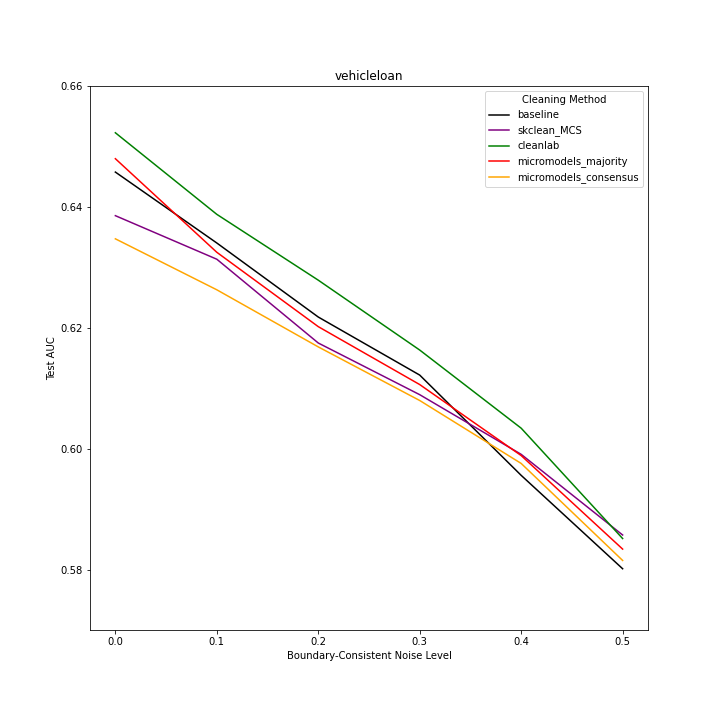}}\\
  \subfloat{\includegraphics[width=0.35\textwidth]{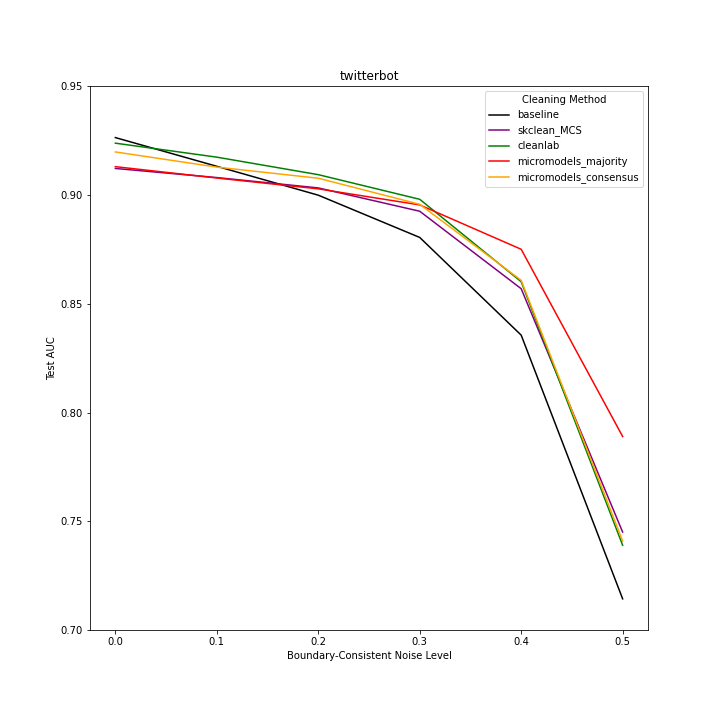}}
\caption{Test ROC-AUC curve following noise removal in training for various datasets using different removal approaches.}
  \label{fig:label_noise_auc}
\end{figure*}

First, in Figure \ref{fig:label_noise_auc} we show the ROC-AUC attained on clean test data by a model trained on data cleaned by the various approaches under consideration. Along the $x$-axis is the amount of noise added to the training data before cleaning, and the $y$-axis shows the resulting AUC. Each cleaning method is colored differently as indicated in the legend. As observed, when there is no noise added, we expect the baseline (no noise removal) to perform the best, and then, as more noise is added, we expect the removal methods to produce the best results. Notable is that CleanLab (in green) typically performs well, nearly always above the baseline; that no method is capable of completely eliminating the effect of noise but are usually good at mitigating it.

In more detail, as expected Baseline is usually (near) the best at zero noise. As noise is added, the other noise removal techniques surpass the baseline - overall performance generally declines as noise is added, though the effective noise removal techniques do not decline as rapidly and in many cases performance is relatively stable with even 10-20\% noise added. As noted earlier, CleanLab performs very well across experiments and particularly at the lower noise levels, it is at higher noise levels that MM-majority and skclean-MCS will take the lead. Exceptions to this are \textit{sparknov} where MM-consensus performs well across the middle range of noise amounts, \textit{vehicleloan} where CL is consistently the best, and \textit{ccfraud} / \textit{fakejob} where MM-majority and skclean-MCS are the best performers across the range of noise levels. Finally \textit{fraudecom} is notable for how little performance varies across noise levels and techniques - this is a difficult dataset, and it is hard for any model to do better than random without feature engineering (as discussed in Section \label{sec:binary}).